\documentclass[runningheads]{llncs}
\usepackage[utf8]{inputenc}

\usepackage{graphicx}

\usepackage[utf8]{inputenc}
\usepackage{amsmath}
\usepackage{bm}
\usepackage{gensymb}
\usepackage{amssymb}
\usepackage{siunitx}
\DeclareSIUnit\px{px}
\DeclareSIUnit\ad{CE}

\usepackage[english]{babel}

\usepackage{microtype}
\usepackage{xspace}
\usepackage{xcolor}
\usepackage{siunitx}
\usepackage[hyphens]{url}
\usepackage{multirow}
\usepackage{booktabs}
\usepackage[pagebackref=true,breaklinks=true,colorlinks,bookmarks=false,citecolor=green!80!black,linkcolor=red!80!black,urlcolor=blue]{hyperref}

\usepackage{cleveref}
\crefname{section}{Sec.}{Sections}
\crefname{figure}{Fig.}{Figure}
\crefname{table}{Tab.}{Table}
\crefname{equation}{Equ.}{Equation}

\usepackage[inline]{enumitem}
\setlist*[enumerate]{label=(\arabic*)}

\usepackage{subcaption}
\usepackage{glossaries}

\newcommand{\onedot}{.\xspace}
\newcommand{\etal}[1]{#1~et~al\onedot}

\newcommand{\eg}{e.\,g.,\xspace}

\newcommand{\cf}{cf\onedot}
\newcommand{\ie}{i.\,e.,\xspace}

\newcommand{\aka}{a.\,k.\,a\onedot}

\begin{document}
\title{The Notary in the Haystack -- %
Countering Class Imbalance in Document Processing with CNNs}
\titlerunning{Countering class imbalance in medieval document sets}
\author{Martin Leipert\inst{1} \and
Georg Vogeler\inst{2} \and
Mathias Seuret\inst{1} \and
Andreas Maier\inst{1} \and \\
Vincent Christlein\inst{1}
}
\authorrunning{Leipert et al.}
\institute{Pattern Recogntition Lab, FAU Erlangen-Nuremberg, Erlangen, Germany\\
	\email{\{firstname.lastname\}@fau.de} \and 
Austrian Centre for Digital Humanities, Universität Graz, Graz, Austria\\ 
		\email{georg.vogeler@uni-graz.at}}
\maketitle              %
\setcounter{footnote}{0}
\begin{abstract}

Notarial instruments are a category of documents. 
A notarial instrument can be distinguished from other documents by its notary
sign, a prominent symbol in the certificate, which also allows to identify the
document's issuer.
Naturally, notarial instruments are underrepresented in regard to other
documents. 
This makes a classification difficult because class imbalance in training data worsens the performance of Convolutional Neural Networks.
In this work, we evaluate different countermeasures for this problem. 
They are applied to a binary classification and a segmentation task on a collection of
medieval documents. 
In classification, notarial instruments are distinguished from other documents,
while the notary sign is separated from the certificate in the segmentation task.
We evaluate different techniques, such as data augmentation, under- and
oversampling, as well as regularizing with focal loss. 
The combination of random minority oversampling and data
augmentation leads to the best performance. 
In segmentation, we evaluate three loss-functions and their combinations,
where only class-weighted dice loss was able to segment the notary sign sufficiently. 

\end{abstract}

\newacronym{NLL}{NLL}{Negative Log Likelihood}
\newacronym{CE}{CE}{Cross Entropy}

\section{Introduction}
\begin{figure}[t]	
	\centering
	\subcaptionbox{Source: Seckau, Austria, 1455, BayHStA München HU Chiemsee
		23.\footnotemark\label{fig:sign}}
		[0.4\textwidth]{
			\includegraphics[height=1.5in]{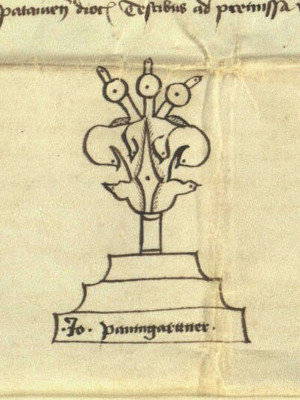}
		}
	\qquad
	\subcaptionbox{Source: Vienna, Austria, 1460, Archiv der Erzdiözese Salzburg, AUR
		2836.\footnotemark\label{fig:notary_full}}
		[0.4\textwidth]{
		\includegraphics[height=1.5in]{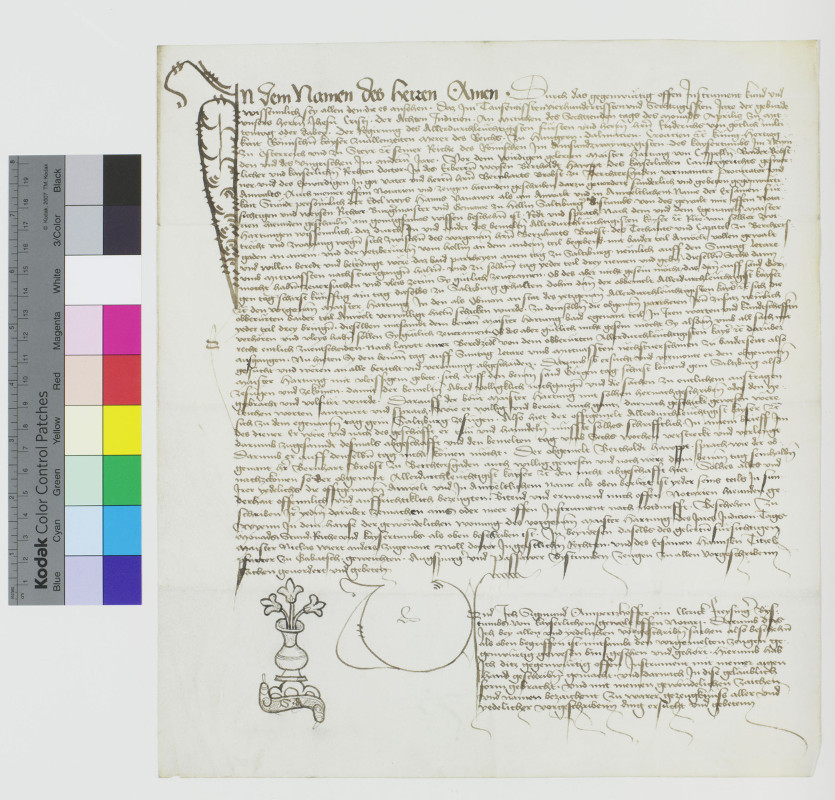}
	}
		\caption{Examples of a notary sign \subref{fig:sign} and a notary document \subref{fig:notary_full}.}
		\label{fig:examples}
\end{figure}

Notarial instruments, the subject of this work, originate from medieval Italy. 
The ones considered in this study are recognizable for the human eye by a characteristic notary sign, see for
example \cref{fig:sign}, which served as signature to identify the issuing
commissioner. 
Around \SI{1200}{\ad} they appeared in today's Germany and Austria. 
The analysis of their spread and development is an open problem in historic research~\cite{Haertel11,Weileder14,Weileder19}. 
Especially as they are one particular source to reconstruct the development of monasteries. 
Compared to the number of medieval documents with seals, notarial instruments are underrepresented. 
Automatic classification and segmentation of these documents, to automatically identify the commissioner, is beneficial. 
Notary signs cover typically a small area of the document~\cite{rueckert2011:RUC}, see
\cref{fig:notary_full}, where the notary sign is visible in the bottom left. In classification tasks, the class notary document is severely underrepresented and the difference to other documents is only a small detail. In segmentation tasks, the area of the \textit{notary sign} is much smaller compared to \textit{text}. So both problems have to deal with severe class imbalance.

In this work, we conduct a comparative study for handwritten document
classification and segmentation. These studies are sparse for the field of class
imbalance~\cite{BUDA2018249:BUD}, which is a major problem of the dataset at hand. 
Therefore, we evaluate different techniques to counter this issue: 
\begin{enumerate*} 
	\item augmentation, including a novel idea of data interpolation, which is easily integratable in
		the case of notarial instruments containing characteristic notary signs;
	\item data oversampling and undersampling;
	\item different losses, such as focal loss~\cite{FocalLoss}; 
	\item lastly, we tried to enhance the performance of segmentation networks by feeding them entire
		regions of the layout more often to let them memorize their characteristics.
\end{enumerate*}

The work is structured as follows. After presenting the related work in \cref{0031-subsec:related-work},
we introduce the used dataset in \cref{0031-sec:mats-and-metds}. 
The networks, tools and loss functions used for the study are presented in
\cref{0031-subsec:nets,0031-subsec:tools,0031-subsec:loss}. 
Afterwards, we give an overview of the used settings in \cref{0031-subsec:settings}. 
The results are described in \cref{0031-subsec:classificiation-results} for classification and
\cref{0031-subsec:segmentation-results} for segmentation and discussed in \cref{0031-sec:discussion}.  
Finally, the paper is concluded in \cref{0031-sec:conclusion}. 

\footnotetext[1]{\url{https://www.monasterium.net/mom/AT-AES/Urkunden/2836/charter}}
\footnotetext[2]{\url{https://www.monasterium.net/mom/DE-BayHStA/HUChiemsee/023/charter}}

\section{Related Work}
\label{0031-subsec:related-work}
\paragraph{Countering class imbalance:} There exist two main strategies to
counter class imbalance in neural network training~\cite{Johnson2019}. 
Data level techniques -- which modify the data fed to the model -- and algorithmic
methods which modify the training algorithm. Usually, data level methods try to achieve class
balance. From those, random minority class oversampling achieved the best
performance in previous studies~\cite{BUDA2018249:BUD}.
In contrast to traditional machine learning, oversampling does not lead to overfitting for
convolutional neural networks (CNNs). 
Another technique is SMOTE~\cite{Wong2016UnderstandingDA}, a method to
interpolate new samples from training data. 
Furthermore, data augmentation can be used to virtually increase the amount of training data, although
the performance under augmentation is always worse than with the equivalent amount of real data.
Algorithmic level methods include class-weighting -- to increase the weight adaptation for
miss-classifications -- or specific loss functions, such as focal
loss~\cite{FocalLoss}. 
The latter puts additional weight on hard to classify
examples. This technique (in combination with the RetinaNet architecture)
outperformed classical detectors significantly.

\paragraph{Document classification and segmentation with CNNs:}
Documents of the same genre have a similar visual structure and region-specific features. This
allows to classify documents via CNNs.
In a comparative study, \etal{Harley}~\cite{Harley:2015} showed that pretrained CNNs (on ImageNet) that took the entire document as input,
outperformed other approaches that used only document parts.
Documents may be processed based on simple, textural features.
For example, \etal{Oyedotun}~\cite{Oyedotun2016} achieved a region identification rate (share of regions assigned to the correct class) of \SI{84}{\percent}
for segmentation with a simple network taking six pixel-wise textural features
as input and producing a three-class output (image, text, background). 
Therefore the input was fed to a single fully-connected
hidden layer. Also specifically developed deep CNNs for documents exist. For example, the
\textit{dhSegment} CNN~\cite{Oliveira2018dhSegmentAG} consists of a contracting
path and an expanding path similar to the well-known
U-Net~\cite{Ronneberger2015:RON}.  
The contracting path is based on the \textit{ResNet50} architecture~\cite{He16a}. 
The approach outperformed previous approaches in border and
ornament extraction of medieval documents.

\section{Materials and Methods}
\label{0031-sec:mats-and-metds}
The used document collection originates from the \textit{Monasterium} project. 
This collection of digitized medieval documents from different European
archives offers descriptions and images of more than \num{650000}
documents~\cite{heinz12monasterium:MON,Vogeler19}.

\begin{figure}[t]
	\centering
	\subcaptionbox{Faded font due to water damage}[0.4\textwidth]{\includegraphics[height = 1.0in]{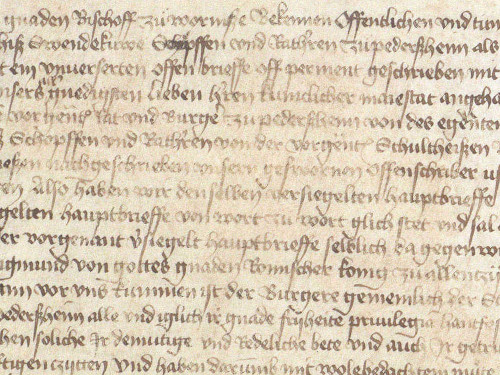}}
	\qquad
	\subcaptionbox{Brown stain}[0.4\textwidth]{\includegraphics[height = 1.0in]{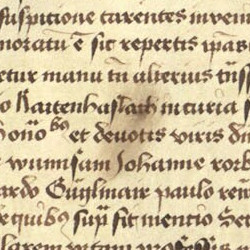}}
	\caption{Examples for varying document appearance.}
	\label{0031-fig:variations}
\end{figure}

Subject of this study is a set of \num{31836} documents, which contains 974 notarial instruments and
various other types. They originate from libraries of Southern Germany and Austria. The notary signs in the documents
are circumscribed by manually drawn polygons. The time range of the documents is from \SIrange{800}{1499}{\ad}, 
where the first notarial instrument in the collection appears shortly after \SI{1200}{\ad}. 
The distributions are not entirely separable, not every notary certificate is clearly identifiable,
\eg due to a missing notary sign. 
The documents are \SI{3000}{\px} wide, but vary in spatial resolution. 
Their real size is between \SI{15 x 20}{\centi\meter} and \SI{40 x
50}{\centi\meter} with font sizes ranging from \SIrange{15}{60}{\px},
the majority around \SI{30}{\px}.  
The size of the notary sign ranges from about \SI{200 x 200}{\px} to \SI{500
x 700}{\px}.
Other varying factors are yellowish paper, low contrast, overexposure during scanning and faded
ink. Stains with low contrast are frequent due to water damage, see for example \cref{0031-fig:variations}.

\subsection{Networks}
\label{0031-subsec:nets}
The classification task is binary, \ie distinguish notarial instruments from
non-notary ones. 
The networks used the full image as input, rescaled and cropped to a size of
\SI{224 x 224}{\px}.
	ResNet50~\cite{He16a} and DenseNet121~\cite{Huang2017densely:HUA}
networks are used for classification, both pretrained on ImageNet. 
Both networks reuse features within the network by skip-connections. 
While ResNet bypasses the
concatenated results from previous layers, DenseNet uses the entire feature vectors of previous
layers within a block. 
DenseNet outperforms ResNet in stability, parameter efficiency, and is less
prone to overfitting~\cite{Huang2017densely:HUA}. 

The segmentation task is a four class problem: background, text, ornaments and
notary signs. As in classification, images are reduced to \SI{224 x 224}{\px}. This
size is sufficient to identify all foreground components (text, ornament and
notary sign).
The patch is segmented with a U-Net~\cite{Ronneberger2015:RON} using 5 layers. 
The U-Net consists of a contracting/encoder and expanding/decoder path. 
Blocks on the same level are interconnected, to reconstruct the output with the
input features on the specific level. 
The U-Net achieves large patch segmentation by an overlap strategy.
Batch normalization~\cite{ioffe2015batch:IOF} is applied in every layer after
the ReLU non-linear functions to stabilize training.

\subsection{Tools}
\label{0031-subsec:tools}
The networks in the classification settings are trained with different settings,
\ie different loss functions, data augmentation and sampling methods.
Adam~\cite{Kingma15} with PyTorch's default parameters ($\beta_1 = 0.9, \beta_2
= 0.999$) is used as optimizer for both the segmentation and the
classification. 
For the classification task, training is performed for 1250 iterations and a batch size of 32. The initial learning rate is $1 \cdot \mathrm{10}^{-3}$, except for settings with focal loss, where it is $5 \cdot \mathrm{10}^{-4}$. To adapt the learning rate, a step decay learning rate scheduler with $\gamma = 0.5$ is used. The step size is set to 250.
In segmentation, training is performed for 60 epochs, where the initial learning
rate is $3 \cdot \mathrm{10}^{-3}$, except for settings with focal loss as part
of the loss function. In the latter cases, the initial learning rate is $1 \cdot \mathrm{10}^{-3}$. Again a step decay learning rate scheduler is applied for learning rate adaption. The decay parameter is set to $\gamma = 0.3$ and the step size is 10.
Additionally in classification, dropout with $p=0.2$ is used for the last fully connected layer
of the respective networks, which empirically improves the
recognition rates.

\paragraph{Image Augmentation:}

The data augmentation strategies are derived from the previously examined
dataset-intrinsic variations. 
Those are contrast, color, lighting, affine transformations, deformations and
damages. 
Three dynamic image augmentation functions are built with the \textit{Albumentations}
framework~\cite{Albumentations2018}. 
The ``weak'' augmentation is designed to mimic the intrinsic variations of the dataset. 
The ``moderate'' augmentation uses higher parameter and probability values. 
The ``heavy'' augmentation also includes additional effects not being part
of the intrinsic data variation.  
These functions are identical for both classification and segmentation.

They are all composed according to a certain pattern, \ie they include a combination of up to seven groups of effects. These groups are a
probabilistic combination of flip, an affine transformation, one of a group of
blurs, one non-linear distortion, a special effect, a brightness-contrast
variation and one of various color modifications as shown in \cref{0031-tab:augmentation_groups}. 

\begin{table}[t]
	\caption{Used groups of augmentation effects. Italic effects only occur for moderate and/or heavy augmentation.}
	\label{0031-tab:augmentation_groups}
	\begin{tabular}{l l} \hline
		Group & Effects \\ \hline
		Flip & Horizontal and vertical flip, 90$^{\circ}$-rotation \\
		Affine Transformation & Random rescale, shift \& rotation \\
		\textit{Blur \& Noise} & \textit{Ordinary \& gaussian noise, ordinary, gaussian}  \\
		&  \textit{motion \& ordinary blur, median filter} \\
		Distortion & Optical \& grid distorsion, elastic transformation \\
		Brightness \& contrast & Random brightness \& contrast adaption. \\
		Color & Histogram equalization (CLAHE), random HSV\footnote{Hue Saturation Value}-shift, \\
		&  RGB-shift, channel shuffle and grey conversion \\
		Special effect & Shadow, \textit{snow or jpg compression} \\ \bottomrule
	\end{tabular}
\end{table}
The ``flip'' and ``affine transformation'' groups simulate different positions, scales and orientations of the document. The flips are used to force the network to learn the features independent of their absolute position. The ``brightness \& variance'' and ``Blur \& noise'' groups mimic aging paper and different acquisition qualities. The ``Distortions'' imitate acquisition variables, as the photographies of the documents are not always plain. The ``Color'' shift simulates different font and paper colors. 
The ``special effects'' group mimics either darker (shadow) or missing areas (snow). 

Each of these groups has an assigned occurrence probability. If the group is drawn, one effect out of the group is chosen -- again according to a probability -- to contribute to the augmentation.
This concept is the same for all three augmentation functions. So in an edge
case, exactly one effect per seven groups is applied.
The occurrence probabilities of the groups and the intensity parameters of the effects are varied, resulting in different augmentation intensities. Each parameter and probability is set such that it doubles from ``weak'' to ``heavy''. As consequence for ``weak'' augmentation in average 1.7 effects were applied, for ``moderate'' 3.4 and for ``heavy'' 4.0.
\eg the group affine transformation occurred with a probability of 0.4 for ``weak'' augmentation and with a probability of 0.7 for ``moderate'' and ``heavy''. The scale limit was 0.1 for ``weak'', 0.15 for ``moderate'' and 0.2 for ``heavy''.

\begin{figure}[t]
	\centering
	\includegraphics[width=6cm]{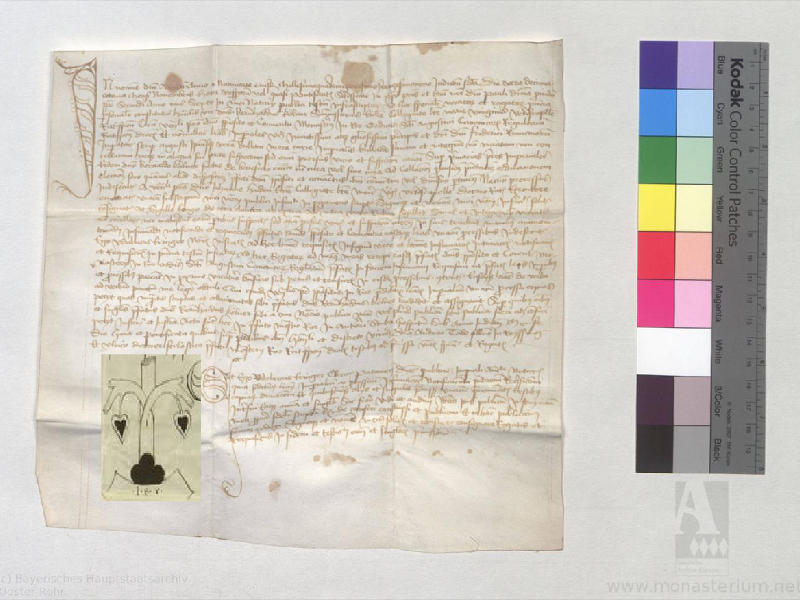}
	\caption[Swapped Notary certificate.]{Notary certificate with the notary sign from another document inserted. Document source:
	BayHStA KU Rohr 327, Sign source: Archiv Erzdiözese Salzburg, D319.\footnotemark}
	\label{0031-fig:swapped_notary_sign}
\end{figure}
\footnotetext[3]{\url{https://www.monasterium.net/mom/AT-AES/Urkunden/2854/charter}}

\paragraph{Notary Symbol Swap:}
This was introduced to allow meaningful interpolation between data. The system
should learn the notary sign as the decisive structural element of
notarial instruments. Therefore, notary signs of other documents are either
inserted at the original sign's position, \cf
\cref{0031-fig:swapped_notary_sign} for an example, or added at an arbitrary
location of the document. 

\paragraph{Random Minority Oversampling:}
Used to balance the dataset, in which non-notary documents occur 32 times more
frequent than notary ones. Hence, the latter ones are oversampled. This is done
by random drawing, such that notary and non-notary signs occur with the same
probability of 0.5. The variance within the notary set is smaller, which is combined with
augmentation to compensate possible overfitting effects.

\paragraph{Random Majority Undersampling:} In contrast to oversampling, a
random subset of the non-notary documents is drawn to balance the number of
samples of notary and non-notary document within the training data.

\paragraph{Meaningful Segment Training:}
In the segmentation task, background and text are largely over-represented in
comparison to notary signs. To compensate this, the U-Net is trained with
segments of an entire ``meaningful'' area of the image (text, ornaments or a
notary sign) by randomly choosing one type of these segments occurring in the
image, this is repeated to fill the image or resized to fill the patch. This is done with a probability of 0.5 alternating with normal notary certificates.

\subsection{Loss Functions}
\label{0031-subsec:loss}
\paragraph{Classification:}
For training, we use the binary cross entropy (BCE) loss. 
As alternative loss function and algorithmic imbalance compensation, focal loss~\cite{FocalLoss} is
used. 
It emphasizes examples with a high error in training. Due to this property, it achieves good
performance for problems suffering from strong class imbalance.  
Focal loss is based on balanced cross entropy and
consists of adding a weighting $\left( 1- p_t\right)^{\gamma}$ and (optionally) a class weighting factor $ \alpha_t $ to the BCE
function. 
\begin{equation}
	\mathrm{FL}\left( p_t \right) = - \alpha_t \left( 1- p_t\right)^{\gamma} \log{\left(p_t\right)}
\end{equation}
where the probability $p_t$ is defined as: 
\begin{equation}
p_t = 
\begin{cases}
p & y = 1 \\
1 - p & \text{otherwise} \\
\end{cases}
\end{equation}

With a high $\gamma \approx 2$, hard examples are weighted stronger and thus are focus of training. 
This setting is used for focal loss in classification. With a low $\gamma \approx 0.5$ the weight of samples with a small deviation is increased. With $\gamma = 0$ the function and its properties are equivalent to BCE loss. 
In the classification tasks $\gamma = 2$ is used.

\paragraph{Segmentation:} We evaluate training of a U-Net using different loss functions:
\begin{enumerate*} 
	\item class-weighted dice loss (weighted by inverse class frequency), 
	\item BCE loss, and 
	\item a two dimensional version of focal loss ($\gamma=1.5$) 
\end{enumerate*}. 
Additionally, we combine different loss functions to evaluate if these combinations improve
training behavior by compromising different optimization goals.

\begin{table}[t]
	\centering
	\caption{Settings for the Training of the DenseNet and the ResNet}
	\label{0031-tab:setups_classification}
	\begin{tabular}{ccccc}
		\toprule
		Setting & Augmentation & Swap/Add Sign & Loss & Under-/Oversampling \\
		\midrule
		1 &  	& 	&  BCE	& 	\\
		2 &  	& 	&  BCE	& +	\\
		3 & o 	& 	&  BCE	& +	\\
		4 & + 	& 	&  BCE	& +	\\
		5 & ++ 	& 	&  BCE	& +	\\
		6 & + &  &  BCE & - \\
		7 & 	& Swap \& Add &  BCE 	& +	\\
		8 & + 	& Swap \& Add &  BCE	& +	\\
		9 & + & Swap &  BCE & + \\
		10 & + &  & Focal & \\
		11 & +	& Swap \& Add	& Focal	& 	 \\
		12 & + & Swap & Focal & + \\
		\bottomrule
	\end{tabular}
\end{table}

\section{Evaluation}

\subsection{Evaluation Protocol}
\label{0031-subsec:settings}
The set of \num{31838} images for the classification was split into \num{21227} training, \num{2653} validation
and 7958 test images. For segmentation, the set of 887 images was split into 496 training, 98
validation and 293 test images.

For the classification task, we created twelve different evaluation settings, see \cref{0031-tab:setups_classification}.
The settings test various augmentation intensities, namely ``weak'' (\textbf{o}), ``moderate'' (\textbf{+}) and ``heavy'' (\textbf{++}) augmentation, swapping of the notary sign
and random adding/swapping of notary signs. 
Focal loss, undersampling (\textbf{-}) and oversampling (\textbf{+}) are compared under moderate augmentation. 
Each setting is executed and trained three times. We sum up all the results and compute common error metrics for a two class-problems:
sensitivity, specificity and the F-value (\aka F1-score).

The first setting serves as baseline where no imbalance counter measures are used. In the second setting, random minority oversampling as most effective counter-measure is introduced. This is further expanded in settings 3, 4, 5, where data augmentation is added to the previous setting and its intensity is steadily increased from ``weak'' to ``heavy'' to further boost the accuracy. The most effective ``moderate'' augmentation is used in setting 6, together with random majority undersampling as a comparison to oversampling. Settings 7 to 9 introduce notary symbol swapping/adding. This method is used as sole augmentation in setting 7. In setting 8, this technique serves as additional augmentation to ``moderate'' augmentation. Setting 9 uses ``moderate'' augmentation and symbol swapping only, to compare if it is more effective than adding. The underlying assumption is, that the correct arrangement of the sign is important for detection and the adding of a symbol at an arbitrary position could possibly reduce recognition accuracy. 
From setting 10 to 12, focal loss is introduced and combined with the best performing ``moderate'' augmentation in setting 10. In setting 11, this augmentation is further combined with swapping and adding of notary signs.
Finally all successful techniques -- random minority oversampling, moderate augmentation and symbol swap -- are combined with focal loss, to examine, if focal loss performance could beat BCE.

\begin{table}[t]
	\centering
	\caption{Settings for the training of the U-Net in the segmentation task.}
	\label{0031-tab:setups_segmentation}
	\begin{tabular}{cccccc}
\toprule
		Setting & Augmentation & Focal & Dice & BCE & Meaningful Segments  \\ 
\midrule
		1 	& 		& 1.0	&		& 		& \\  
		2 	& o		& 1.0 	&		& 		& \\   
		3 	& +		& 1.0 	&		& 		& \\  
		4 	& ++	& 1.0 	&		& 		& \\ 
		5 	& +		&	 	& 1.0	& 		& \\  
		6 	& +		&	 	&		& 1.0	& \\
		7 	& +		&	 	& 0.5	& 0.5	& \\
		8 	& +		& 0.5	&		& 0.5	& \\    
		9 	& +		& 0.5 	& 0.5	& 		& \\    
		10 	& +		& 0.33	& 0.33	& 0.33	& \\   
		11 	& +		&  		&		& 1.0		& + \\  
		12 	& +		&  		& 1.0	& 		& + \\  
		13 	& +		& 1.0	&		& 	& + \\  
\bottomrule	
\end{tabular}
\end{table}

For the segmentation experiments, the main variation in the settings
(\cref{0031-tab:setups_segmentation}) is the use of segments and the combination of different loss functions. In the settings 1 to 4 with focal loss, it is tested if augmentation improves the result. In settings 5 to 10 the most effective combination of loss functions is tested. Settings 11 to 13 check if meaningful segments improve the result.
The segmentation results are evaluated by means of the Intersection over Union (IoU) 
\begin{equation}
	\text{IoU} = \frac{\left| A \cap B \right|}{\left| A \cup B \right|}
\end{equation}
of the labels in the ground truth  $A$ and in the networks' prediction $B$.

\begin{table}[t]
\caption{Classification results for ResNet and DenseNet.}
\centering
\begin{subtable}{.5\textwidth}
\centering
    \begin{tabular}{c c c c}
        \toprule
        Setting & Sensitivity & Specificity & F-Value \\
        \midrule
            1  & \textbf{100} & 0 & 0 \\
            2  & 92.2 & 82.4 & 87.0 \\
            3  & 97.4 & 93.4 & 95.3 \\
            4  & 96.5 & \textbf{94.8} & 95.7 \\
            5  & 95.4 & 94.5 & 94.9 \\
            6  & 94.9 & 94.7 & 94.8 \\
            7  & 96.9 & 84.0 & 89.9 \\
            8  & 97.1 & 92.6 & 94.8 \\
            9  & 96.4 & 93.7 & 95.0 \\
            10 & 94.5 & 51.2 & 66.4 \\
            11 & 94.4 & 89.6 & 91.9 \\
            12 & 97.8 & 94.0 & \textbf{95.9} \\
        \bottomrule
    \end{tabular}
\caption{ResNet}
\label{0031-tab:result_table_resnet}
\end{subtable}%
\begin{subtable}{.5\textwidth}
\centering
    \begin{tabular}{c c c c}
        \toprule
        Setting & Sensitivity & Specificity & F-Value \\
        \midrule
            1  & 90.2 & 56.5 & 69.5 \\
            2  & 93.3 & 87.0 & 90.0 \\
            3  & 98.5 & 95.0 & 96.8 \\
            4  & 97.3 & 94.9 & 96.1 \\
            5  & 97.3 & 94.7 & 96.0 \\
            6  & 96.4 & 96.9 & 96.6 \\
            7  & 97.2 & 89.7 & 93.3 \\
            8  & 98.1 & 94.2 & 96.1 \\
			9  & 98.4 & \textbf{95.3} & \textbf{96.9} \\
            10 & 94.4 & 89.9 & 92.1 \\
            11 & 95.4 & 88.3 & 91.8 \\
            12 & \textbf{99.0} & 94.8 & \textbf{96.9} \\
        \bottomrule
    \end{tabular}
\caption{DenseNet}
\label{0031-tab:result_table_densenet}
\end{subtable}
\end{table}

\subsection{Classification Results}
\label{0031-subsec:classificiation-results}
The classification results for the different settings are given in
\cref{0031-tab:result_table_resnet} and \cref{0031-tab:result_table_densenet} for ResNet and
DenseNet, respectively. 
Ignoring class imbalance (setting 1) leads to the worst results for both networks, the ResNet model
even fails to correctly classify any notarial instrument.
Using oversampling without data augmentation (setting 2) gives the second worst results. 
When combining oversampling with data augmentation (setting 3--6,8,9) the results increase
significantly, but start to decrease with ``heavy'' augmentation (setting 5).
The best results are achieved using the DenseNet with a ``moderate'' augmentation function (setting 3,9) and for the combination of moderate data augmentation, focal loss and oversampling (setting 12).

Swapping the notary sign (setting 7--9,11,12) does neither drastically improve
nor worsens the results.
When using only notary sign swapping \& adding without data augmentation (setting 7), the results
improve compared to pure oversampling (setting 2). 
For DenseNet, the combination of notary sign swapping with data augmentation and oversampling (setting 9) achieves the best F-measure. 
Notary sign swapping worsens the result slightly for ResNet when
applied in combination with ``moderate'' data augmentation, \ie comparing setting 8 with setting 4. 

Comparing undersampling (setting 6) with oversampling (setting 4), the results
are inconclusive, oversampling works slightly better for ResNet and the other
way around for DenseNet. 
Focal loss (setting 10--12) is also able to compensate data imbalance, but does not achieve the same
recognition rates when applied without oversampling (10,11). 
It performs significantly worse than oversampling on moderately augmented data (setting 10 vs.\ setting 4). When adding swapping to oversampling, the accuracy increases for ResNet but does not improve for
DenseNet.

In nearly all settings (apart from setting 11), DenseNet outperforms the corresponding ResNet
setting.  
In particular, all DenseNet settings with over- or undersampling in combination with augmentation
perform better than the best ResNet setting.

\begin{table}[t]
	\centering
	\caption{Intersection over Union (IoU) for different area types.}
	\label{0031-tab:segmentation-results}
	\begin{tabular}{c c c c c c}
		\toprule
		Setting & Background & Text & Ornament & Notary Sign & mean
		IoU\\
		\midrule
		1 	& 0.8820 & 0.8848 & 0.0000 & 0.0022 & 0.5896\\
		2   & 0.8639 & 0.8711 & 0.0000 & 0.0000 & 0.5783\\
		3 	& \textbf{0.8931} & 0.8797 & 0.0000 & 0.0016 & 0.5915\\
		4	& 0.8694 & 0.8544 & 0.0000 & 0.0041 & 0.5760\\
		5 	& 0.8791 & 0.8741 & 0.0000 & 0.0603 & 0.6045\\
		6	& 0.8507 & 0.8503 & \textbf{0.0049} & 0.2902 & 0.6637\\
		7 	& 0.8857 & \textbf{0.8833} & 0.0000 & \textbf{0.4101} & \textbf{0.7264}\\
		8 	& 0.8833 & 0.8622 & 0.0000 & 0.0000 & 0.5818\\
		9 	& 0.8818 & 0.8747 & 0.0000 & 0.0003 & 0.5856\\
		10 	& 0.8860 & 0.8705 & 0.0000 & 0.0001 & 0.5855\\
		11 	& 0.8255 & 0.7995 & 0.0000 & 0.0007 & 0.5419\\
		12 	& 0.7358 & 0.7977 & 0.0027 & 0.1942 & 0.5759\\
		13	& 0.8494 & 0.8237 & 0.0000 & 0.0045 & 0.5592\\
		 \midrule
		 \textbf{Area [\%]} & \textbf{63.95} & \textbf{34.28} & \textbf{0.07} &
		 \textbf{1.70} \\
		\bottomrule
	\end{tabular}
\end{table}

\subsection{Results of the Segmentation}
\label{0031-subsec:segmentation-results}
\Cref{0031-tab:segmentation-results} gives the results for all settings which were tried in this study.
See \Cref{0031-tab:setups_segmentation} for individual details about them.

\begin{figure}[t]
	\centering
	\subcaptionbox{Without Augmentation}{\includegraphics[height = 1.5in]{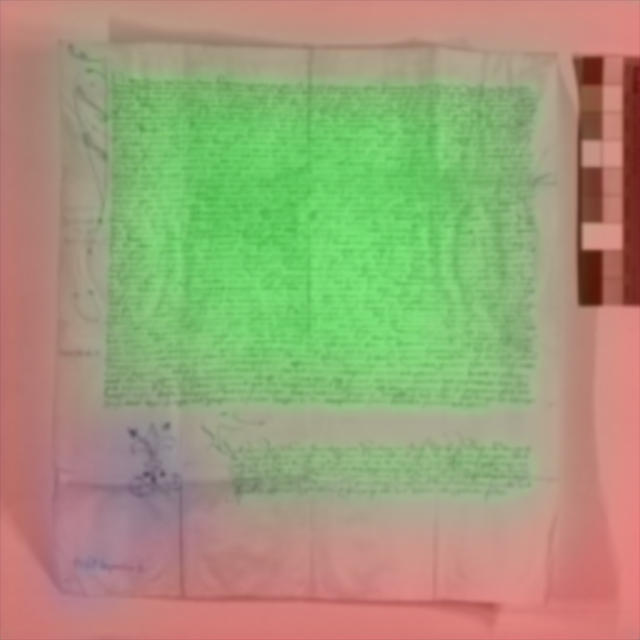}}
	\subcaptionbox{Weak Augmentation}{\includegraphics[height = 1.5in]{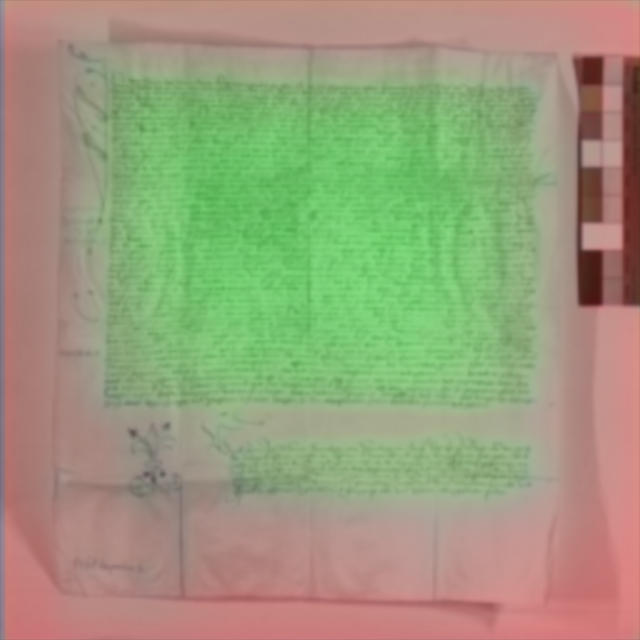}}
	\subcaptionbox{Moderate Augmentation}{\includegraphics[height = 1.5in]{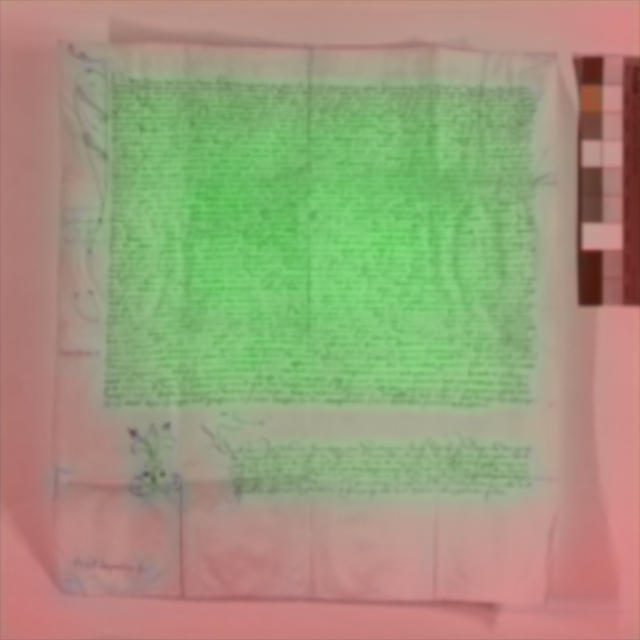}} \\
	\subcaptionbox{Heavy Augmentation}{\includegraphics[height = 1.5in]{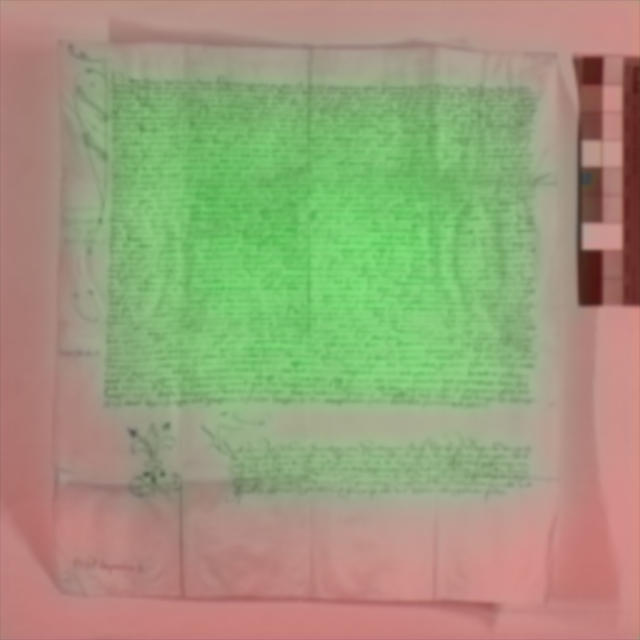}}
	\subcaptionbox{Probability of Text}{\includegraphics[height=1.5in]{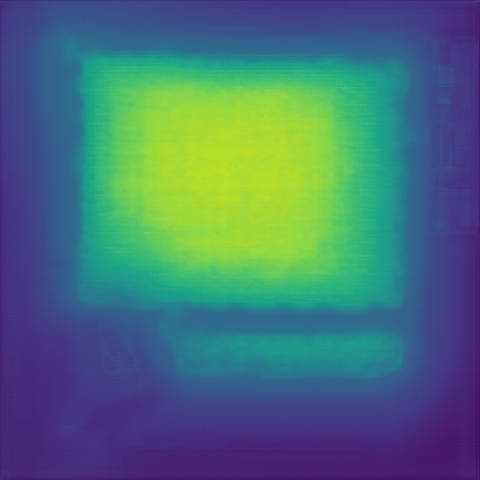}}
	\subcaptionbox{Probability of Notary Sign}{\includegraphics[height=1.5in]{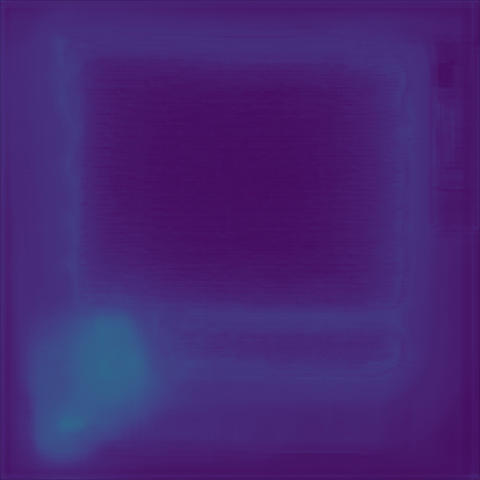}}
	\caption{Comparison of the segmentation results with focal loss and different augmentations. In the  plots \textit{a}, \textit{b}, \textit{c} \& \textit{d} red color indicates class ``background'', green ``text'' and blue ``notary sign''. Source: BayHStA München KU Niederaltaich 951.}
	\label{0031-fig:focal-loss-augmentations}
\end{figure}

\paragraph{Focal Loss with Different Augmentations (setting 1--4):}
\Cref{0031-fig:focal-loss-augmentations} shows the results of focal loss for
different augmentation levels as colored areas. At all augmentation levels, the network
trained with focal loss is able to distinguish well text and background. 
However, it fails entirely to segment ornaments or notary signs, see also
\cref{0031-tab:segmentation-results} setting 1--4. Additionally, the IoU declines
slightly with stronger augmentation as the best result is achieved without augmentation. 

\begin{figure}[t]
	\centering
	\subcaptionbox{BCE loss}{\includegraphics[height = 1.5in]{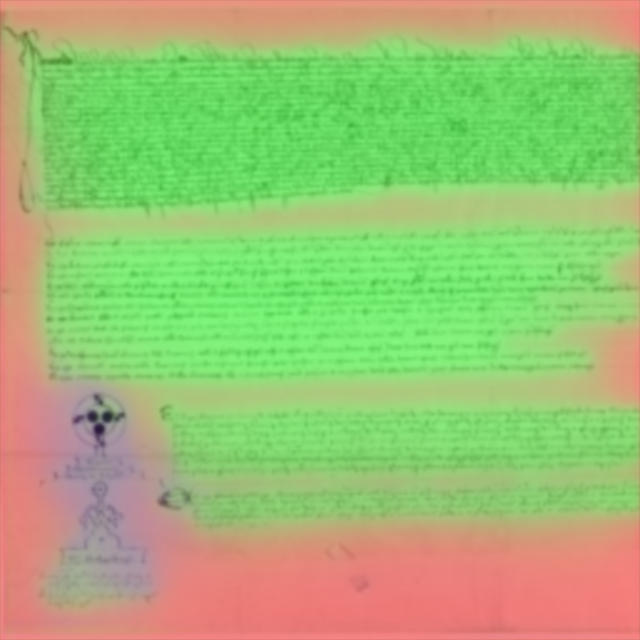}}
	\subcaptionbox{Dice loss\label{0031-fig:moderate-aug-segmentation_b}}{\includegraphics[height = 1.5in]{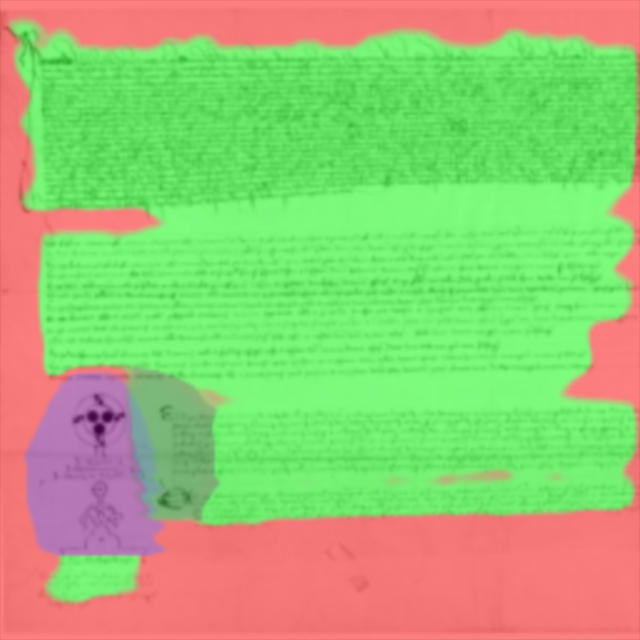}}
	\subcaptionbox{Focal Loss}{\includegraphics[height = 1.5in]{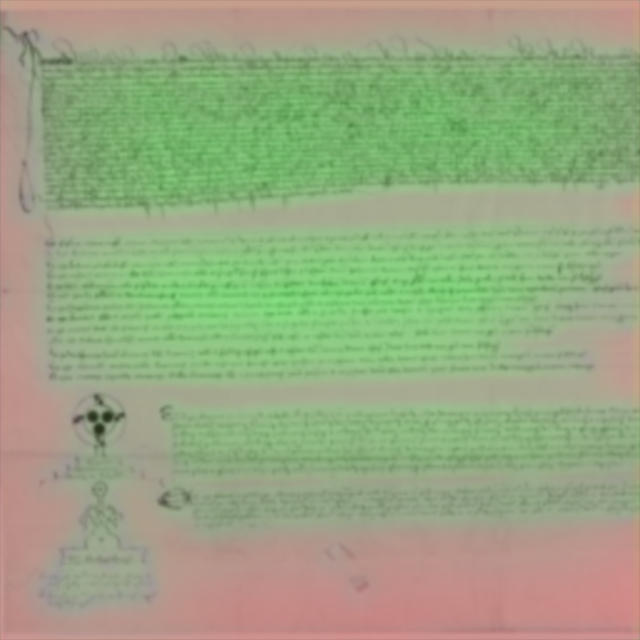}}
	\caption{Comparison of the segmentation results with moderate augmentation and
	different loss functions (color sheme is identical to \cref{0031-fig:focal-loss-augmentations}). Source: Archiv der Erzdiözese Salzburg, D467.}
	\label{0031-fig:moderate-aug-segmentation}
\end{figure}

\paragraph{Moderate Augmentation and Different Loss Functions (setting 3,5,6)} 
A qualitative comparison of the different losses in combination with moderate augmentation can be found in \cref{0031-fig:moderate-aug-segmentation}, the
color scheme is similar to the focal loss example.
Dice loss is well able to segment notary signs, although it segments a larger area around the
sign. Also the text is segmented mainly correctly, with comparable results to BCE and focal loss.
Additionally, Dice loss delivers the most confident result. The only drawback
are artifacts, such as the lower area of the notary sign in
\cref{0031-fig:moderate-aug-segmentation_b} that is classified as ``text''.
Also non-existing ``notary signs'' appear several times in the lower left corner of documents. 
BCE segments text and background relatively
accurately (setting 5), even better than Dice (setting 6), \cf \cref{0031-tab:segmentation-results}. 
The segmentation for the text is especially tight. 
However it fails, similar to focal loss, to segment notary signs.
Focal loss behaves comparable to BCE, with the only difference that BCE segments the notary signs slightly better, however, still insufficiently.

\begin{figure}[t]
	\centering
	\subcaptionbox{BCE and Dice Loss}{\includegraphics[height = 1.15in]{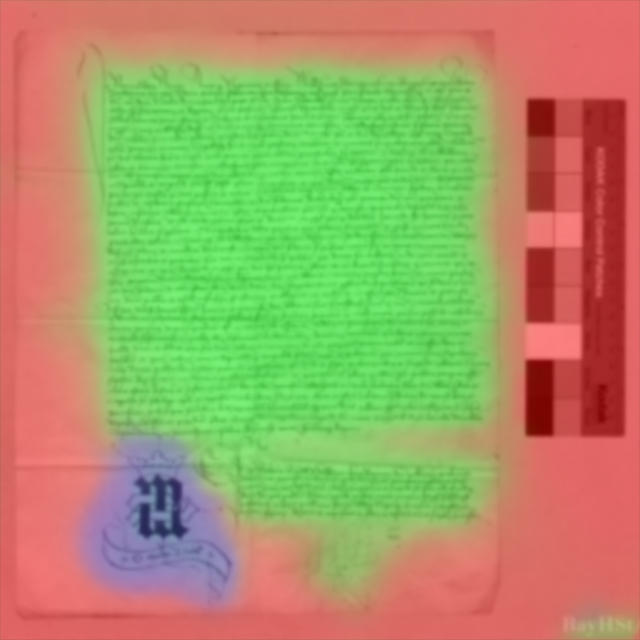}}
	\subcaptionbox{BCE and Focal Loss}{\includegraphics[height = 1.15in]{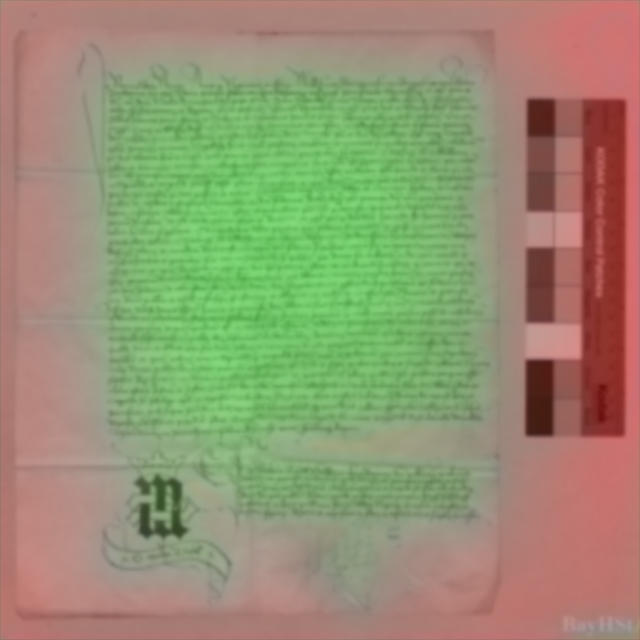}}
	\subcaptionbox{Dice and Focal Loss}{\includegraphics[height = 1.15in]{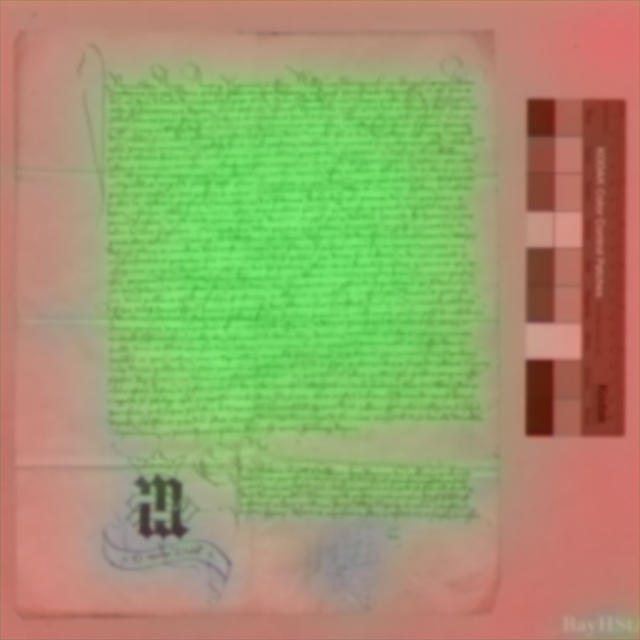}}
	\subcaptionbox{BCE, Dice and Focal Loss}{\includegraphics[height = 1.15in]{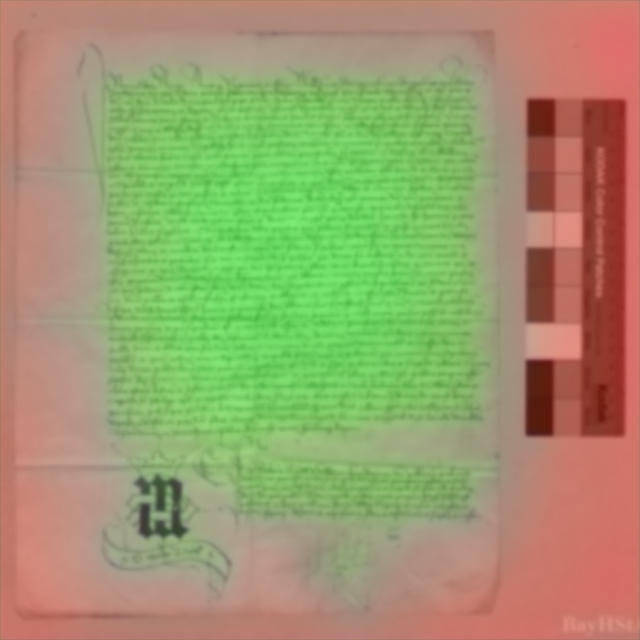}}
	\caption{Comparison of the segmentation results with moderate augmentation and different loss function combinations (color sheme is identical to \cref{0031-fig:focal-loss-augmentations}). Source: BayHStA München KU Aldersbach 877}
	\label{0031-fig:combined-loss-segmentation}
\end{figure}

\paragraph{Combined Loss Functions (Setting 7--10):}
\Cref{0031-fig:combined-loss-segmentation} shows the results of the combined loss functions. 
Dice and BCE (setting 7) are a good combination providing overall accurate boundaries and
segmented the notary sign clearly and overall achieve the best results in \cref{0031-tab:segmentation-results}. All combinations with focal loss perform significantly worse.

\begin{figure}[t]
	\centering
	\subcaptionbox{BCE Loss}{\includegraphics[height = 1.25in]{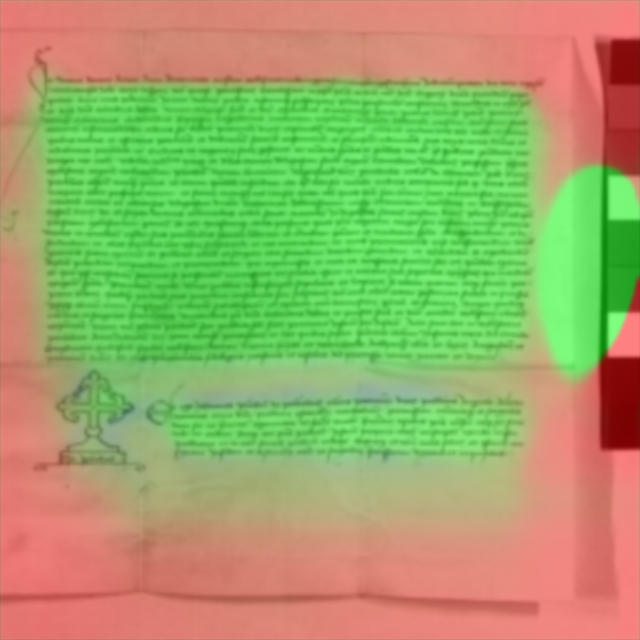}}
	\subcaptionbox{Dice Loss}{\includegraphics[height = 1.25in]{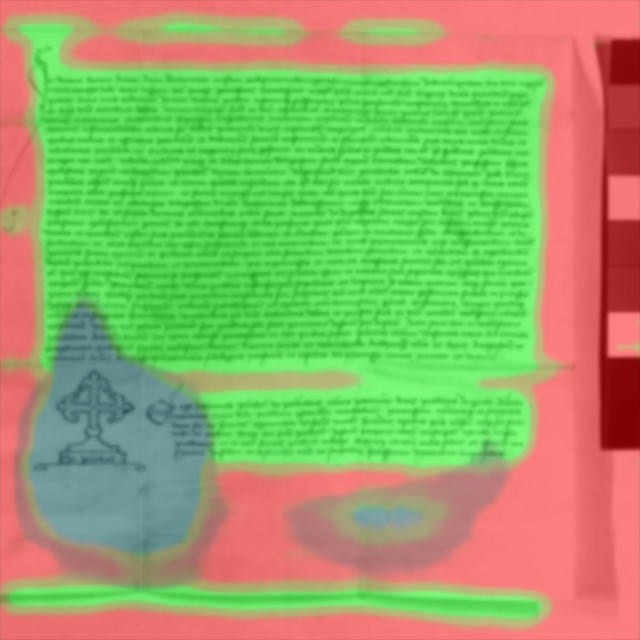}}
	\subcaptionbox{Focal Loss}{\includegraphics[height = 1.25in]{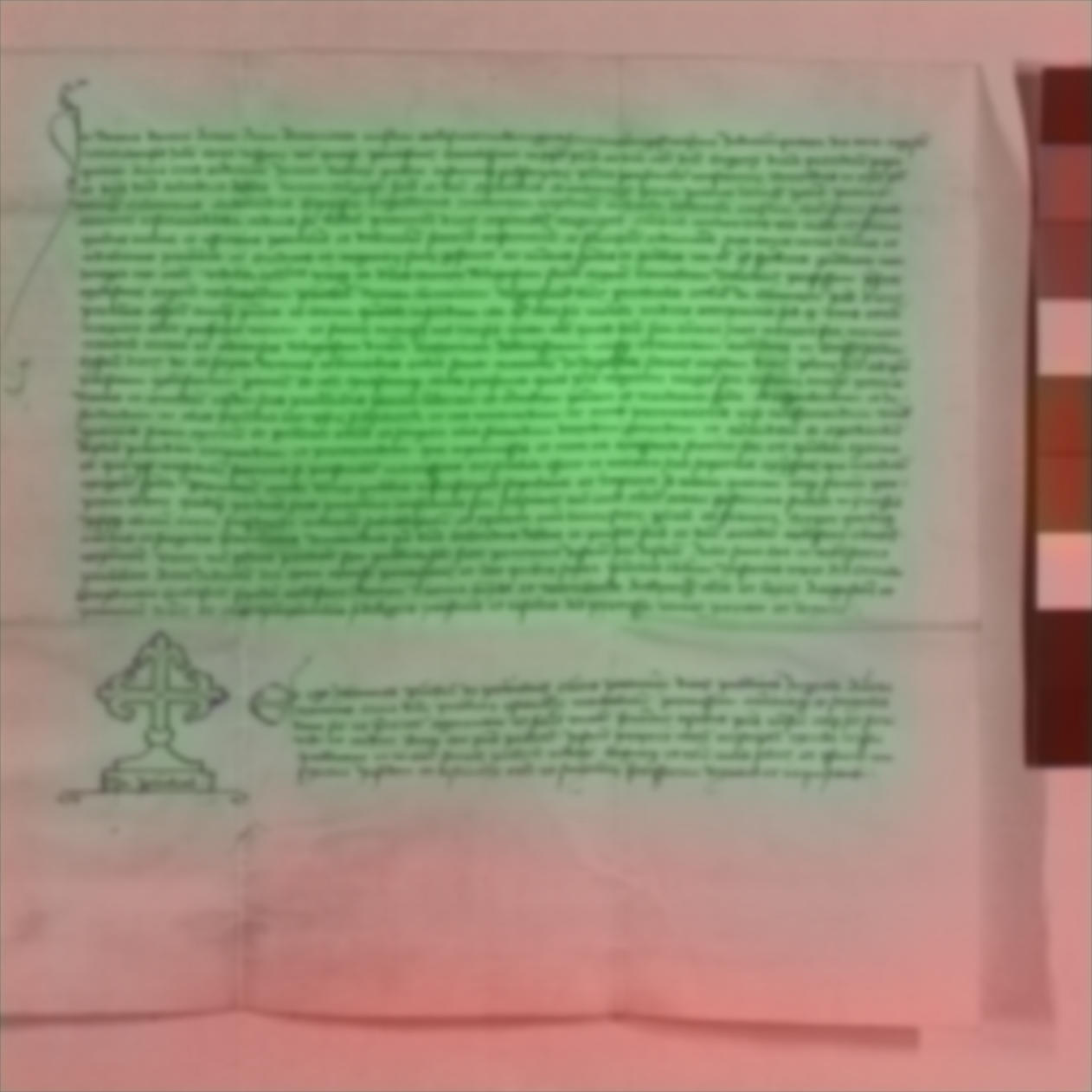}}
	\caption{Results of segmentation trained with snippets (color sheme is identical to \cref{0031-fig:focal-loss-augmentations}). Source: BayHStA München KU Niederaltaich 2526}
	\label{0031-fig:snippet-segmentation}
\end{figure}

\paragraph{Training with meaningful region snippets (Setting 11--13)}
The region snippets deteriorate the result's quality in terms of confidence and
accuracy. This is valid for both repeating and scaling, where the first performs better in experiments than the latter. This is also demonstrated in \cref{0031-fig:snippet-segmentation}. 
It leads to artifacts and area miss-classifications. For any used loss function,
the prediction is worse than with the corresponding network trained without this
training scheme. 

\subsection{Discussion}
\label{0031-sec:discussion}
\paragraph{Classification} Countering class imbalance is important, data
augmentation, focal loss, under- and oversampling improve the results
significantly in classification. However, focal loss performs worse as
single countermeasure. 
In combination with over-/undersampling, focal loss performs slightly better than
pure BCE and decreases missclassification rate of the majority class, with
the price of a increased missclassification rate for the minority class.
It outperforms other settings, when it is combined with data augmentation, notary sign
swapping and oversamplinng.

Data interpolation with swapping and adding notary signs is a valid counter strategy for class imbalance. 
The boost is however weak in comparison to data augmentation. 
Nevertheless, when applied in combination with data augmentation, it works as best boost for accuracy. 

Data augmentation clearly improves the prediction. 
Despite this, there is a limit for that effect when the data quality starts to
diminish (shown with ``heavy'' augmentation). 
Improvements by augmenting data contribute up to a \SI{6.3}{\percent} increase
(ResNet) and \SI{6.7}{\percent} increase (DenseNet) in F-value compared to plain
oversampling.

\paragraph{Segmentation} 
For segmentation, the best results are found with class-weighted Dice loss in combination with Binary Cross Entropy. Pure Dice loss performs also well. The results with focal loss and BCE are less confident and often fail to segment the notary sign. 
Focal loss is used with a focus on hard-to-classify examples. The notary sign is
not recognized well by the network. One reason is probably that the manual
segmentation lead to non-distinct boundaries, causing areas assigned to the wrong class to be treated as hard examples. 
The BCE loss function does not add class-weighting, in contrast to the used class-weighted Dice loss. 
We assume that with accurately estimated examples and clearly separable
distributions, focal loss would perform better in segmentation since focal loss
would compensate the missing class weight. Another factor, which reduces the
training success with focal loss, is that some documents have a very bad data quality, with a strongly faded text. 
The desired effect of an improved behavior for combined loss functions occurs with Dice and BCE loss. All other combinations -- all including focal loss -- deteriorate the results quality. The reasons are probably the same as for plain focal loss. 

Feeding segments during training led to no improvements. When using scaled areas scaling to fill the image, the size of fed areas varies largely, as notary signs are usually much smaller than text area. This probably leads to inconsistency in the kernels, to decreased prediction quality and confidence, and to frequent artifacts. Repeating them probably conflicts with the overlap strategy of UNet and deteriorates region border recognition.
An implementation where the areas would be of normal scale, but notary
signs are fed more frequently without repeating, would probably lead to better results.

\section{Conclusion}
\label{0031-sec:conclusion}
An optimal counter strategy for class-imbalance in two class classification problems is found with
oversampling, notary sign swapping and moderate augmentation. 
Focal loss also works as imbalance counter. The results are comparable as with
BCE when we use the same strategies as the best method, otherwise it performs
worse.  We identify the interpolation of data by swapping notary signs as a
valid technique for data augmentation. However note that normal data
augmentation is more efficient.

In segmentation, Dice loss with class weighting works well for data with high class imbalance. In
combination with BCE its result even improves. This delivers clear segmentation results and sharp
borders. Additionally, a good notary sign segmentation is achieved. Augmentation in case of class
imbalance leads to an improvement for the focal loss function. 
Focal loss fails to segment the notary sign and is inappropriate for this
application and data quality. Apparently, focal loss has problems with the deviations within the
data, which are not always that clearly separable. Also feeding meaningful
segments does not lead to improved behavior, presumably due to too large differences between segment sizes and conflicts with the UNet's overlap strategy.

For future work, we would like to vary the imbalance to see the effect on the
current strategies and also improve the notary sign swapping technique using
noise and other transformations. 

\bibliographystyle{splncs04}
\bibliography{paper}
\end{document}